\documentclass[10pt,twocolumn,letterpaper]{article}

\usepackage[pagenumbers]{wacv} 
\usepackage{graphicx}
\usepackage{multirow}
\usepackage{tikz}

\newcommand{\tikzcircle}[2][black,fill=black]{\tikz[baseline=-0.5ex]\draw[#1,radius=#2] (0,0) circle ;}

\usepackage[ruled,vlined]{algorithm2e}
\usepackage{amsmath,amssymb}
\usepackage[table]{xcolor}
\usepackage{adjustbox}
\usepackage{tabularx}
\usepackage{booktabs,tabularx,ragged2e,xcolor,colortbl}

\definecolor{wacvblue}{rgb}{0.21,0.49,0.74}
\usepackage[pagebackref,breaklinks,colorlinks,allcolors=wacvblue]{hyperref}
\usepackage[capitalize]{cleveref}
\def\wacvPaperID{1848} 
\def\confName{WACV}
\def\confYear{2026}

\title{Guiding What \textit{Not} to Generate:\\ Automated Negative Prompting for Text-Image Alignment}

\author{
  Sangha Park$^1$ \hspace{0.2em}
  Eunji Kim$^2$ \hspace{0.2em}
  Yeongtak Oh$^1$ \hspace{0.2em}
  Jooyoung Choi$^1$ \hspace{0.2em}
  Sungroh Yoon$^{1,3}$\textsuperscript{†}
  \vspace{0.3em} \\
  $^1$Department of Electrical and Computer Engineering, Seoul National University, \hspace{0.2em}  $^2$Amazon \\
  $^3$IPAI, AIIS, ASRI, INMC, and ISRC, Seoul National University \\
}

\begin{document}
\maketitle
\begin{abstract}
Despite substantial progress in text–to–image generation, achieving precise text–image alignment remains challenging, particularly for prompts with rich compositional structure or imaginative elements. To address this, we introduce \textbf{Negative Prompting for Image Correction (NPC)}, an automated pipeline that improves alignment by identifying and applying negative prompts that suppress unintended content. We begin by analyzing cross-attention patterns to explain why both targeted negatives—those directly tied to the prompt’s alignment error—and untargeted negatives—tokens unrelated to the prompt but present in the generated image—can enhance alignment. To discover useful negatives, NPC generates candidate prompts using a verifier–captioner–proposer framework and ranks them with a salient text-space score, enabling effective selection without requiring additional image synthesis. On GenEval++ and Imagine-Bench, NPC outperforms strong baselines, achieving 0.571 vs. 0.371 on GenEval++ and the best overall performance on Imagine-Bench.
By guiding what not to generate, NPC provides a principled, fully automated route to stronger text–image alignment in diffusion models. Code is released at \url{https://github.com/wiarae/NPC}.
\end{abstract}
    
\let\thefootnote\relax\footnotetext{†Corresponding author}
\section{Introduction}
Text-to-image (T2I) generation has surged with the proliferation of large-scale diffusion models~\cite{peebles2023scalable, hurst2024gpt}—including Stable Diffusion~\cite{rombach2022high, podellsdxl, esser2024scaling}, DALL-E 3~\cite{betker2023improving}, and FLUX~\cite{black-forest-labs-no-date}—which have revolutionized content creation by enabling diverse, photorealistic images from natural language. Despite these advances, models still frequently fail to satisfy prompt alignment in practice, particularly for prompts with rich compositional structure (multiple objects, attributes, relations, and counts)~\cite{hu2024ella, li2024genai} and surreal instructions (e.g., a square soccer ball)~\cite{huberman2025image, park2024rare}. 

To improve alignment, recent work has explored self-correction via iterative critique-and-revision~\cite{chen2025t2i}, fine-tuning with large-scale synthetic alignment data~\cite{ye2025echo}, and chain-of-thought–based prompt rewriting~\cite{wang2025promptenhancer}. These methods primarily focus on \emph{positive guidance} in the classifier-free guidance (CFG) sense—that is, strengthening conditioning on the desired content.


\begin{figure*}[t]
  \begin{center}
\includegraphics[width=\textwidth]{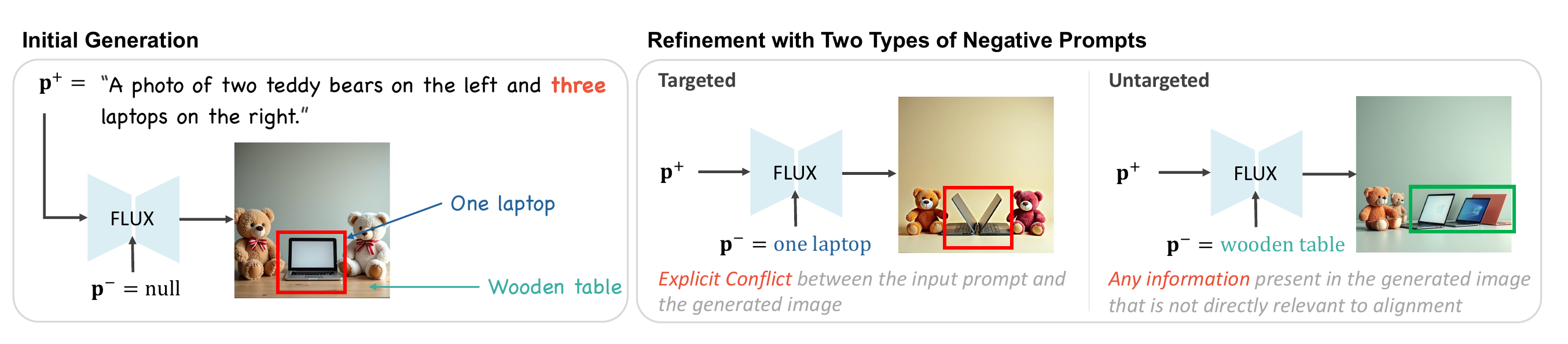}
  \end{center}
  \vskip -0.25in
   \caption{Example of two types of negative prompts used for CFG-based negative guidance: targeted negatives (directly tied to alignment errors) and untargeted negatives (not explicitly related to alignment but appearing in the generated image). Although targeted negatives directly address the error, they do not always ensure alignment, whereas untargeted negatives may still succeed in improving it.}
   \label{fig:non_targeted}
\end{figure*}
Complementing these positive-guidance approaches, a parallel strategy improves alignment by explicitly telling the model what not to generate—suppressing or removing interfering content. Prior efforts typically exhibit two characteristics: (i) layout-centric pipelines~\cite{lian2024llmgrounded,wu2024self}, which constrain element placement by pruning objects, attributes, and spatial relations, but often at the cost of realism due to rigid or imperfect layouts~\cite{park2025raretofrequent,zhang2024realcompo}; and (ii) methods addressing \textbf{targeted} misalignments~\cite{yang2024mastering,wu2024self}, suppressing elements that contradict the positive prompt (e.g., “one laptop" when three laptops are requested; see~\Cref{fig:non_targeted}).

Instead of relying on explicit layouts, \emph{negative guidance} (in the CFG sense) can discourage the generation of particular content, alleviating realism issues. From this perspective, beyond such targeted misalignments, there exist other candidates for negative prompts—details that appear in the image generated from the positive prompt but do not directly harm alignment (which we refer to as \textbf{untargeted} negatives, e.g., a “wooden table”)—whose impact on alignment, to our knowledge, has not been systematically analyzed. To address the lack of systematic analysis of untargeted negatives and of their combination with targeted negatives, we propose a pipeline to \textbf{discover effective negative prompts that enhance alignment} when used as negative guidance.

Moving beyond conventionally used, quality-oriented negatives (e.g., “low quality,” “blurry”), we consider a broader class of negative prompts consisting of both \emph{targeted} and \emph{untargeted} variants. In \Cref{sec:understanding}, we analyze their effect on alignment from a transformer-attention perspective. We compute image–text cross-attention from the denoising latents to the positive-prompt tokens across diffusion timesteps and show that introducing either type of negative guidance increases the attention allocated to salient tokens in the positive prompt. However, identifying which negatives will be effective for a given prompt–image pair remains nontrivial, making naïve trial-and-error regeneration expensive and motivating the need for an automated selection mechanism.

Building on this, we introduce \textbf{NPC}, an automated pipeline that minimizes generation trials: (i) it generates multiple candidate negative prompts via a verifier–captioner–proposer framework, and (ii) uses a \emph{salient score} to select the most effective negative prompt, achieving alignment with only a few on-image trials (see \Cref{fig:npc}). Concretely, NPC works in four steps. First, the \emph{verifier} checks an image created from the original prompt; if the image is misaligned, it points out specific failure reasons, which we treat as \textbf{targeted} negatives. Second, the \emph{captioner} describes the image in natural language, and these captions provide \textbf{untargeted} negatives. Third, the \emph{proposer} combines the \emph{verifier}’s failure reasons with the captions to form several negative-prompt candidates. Finally, we score these candidates with a \emph{salient score}—a text-based measure of how strongly each negative prompt is expected to contribute to improving alignment. Prioritizing the highest-scored candidate identifies an effective negative and reduces the number of regeneration attempts.

We validate NPC on two challenging benchmarks —GenEval++~\cite{ye2025echo}, which emphasizes compositional control, and Imagine-Bench~\cite{ye2025echo}, which evaluates surreal transformations while preserving identity. Across both benchmarks, our approach outperforms strong contemporary T2I baselines. On GenEval++, NPC attains an overall score of 0.571, surpassing the strongest baseline Bagel~\cite{deng2025emerging}, which scores 0.371, by 0.200—a relative gain of about 54\%. On Imagine-Bench, NPC likewise achieves the best overall performance among baselines. Ablation studies further confirm the effectiveness of NPC.
 Our contributions are:
\begin{enumerate}
    \item We expand the scope of negative prompting beyond conventional usage and provide an attention-based analysis of how these negative prompts improve alignment.
    \item We propose an automated pipeline for the generation and selection of negative prompts, thereby reducing manual prompting and human intervention.
    \item We demonstrate superior performance on two challenging benchmarks, consistently exceeding state-of-the-art text-to-image models.
    
\end{enumerate}

\section{Related work}

\subsection{Text-to-Image Diffusion Models}
Diffusion models~\cite{ddpm,sohl2015deep} generate data from noise and have become a strong image-generation paradigm thanks to stable training and scalability~\cite{esser2024scaling,rombach2022high}. GLIDE~\cite{nichol2021glide} first extended diffusion to text-to-image using CLIP embeddings~\cite{radford2021learning} and classifier-free guidance~\cite{cfg}. Imagen~\cite{imagen} underscored the value of richer text encoders (T5~\cite{chung2024scaling}), and recent systems like DALL-E 3~\cite{betker2023improving} and Stable Diffusion 3~\cite{esser2024scaling} show that MLLM-synthesized captions~\cite{gpt4} help with long, complex prompts.

Recent work increasingly pursues \emph{unified multimodal generators} that couple understanding and generation within a single architecture (e.g., Show-o, Transfusion, BLIP3-o, Bagel, OmniGen2, Janus-Pro)~\cite{xie2025showo,zhou2024transfusion,chen2025blip3,deng2025emerging,wu2025omnigen2,chen2025janus}. Whereas most efforts enrich the prompt or strengthen instruction following, we take a complementary direction: we design \emph{effective negative prompts} that selectively suppress off-target content while preserving the intended semantics.

\subsection{Text-Image Alignment}
\paragraph{Alignment methods}
Recent work has improved compositional T2I generation across attribute binding, object relations, numeracy, and complex prompts~\cite{liu2022compositional,guo2024initno,wang2023tokencompose,zhang2024realcompo}. Approaches refine cross-attention to strengthen text alignment~\cite{chefer2023attend,rassin2024linguistic}, impose or predict layouts to control relations~\cite{xie2023boxdiff,wang2024instancediffusion,li2023gligen,lian2024llmgrounded,yang2024mastering}, and leverage MLLMs or human/reward feedback for compositionality and fidelity~\cite{singh2023divide,jiang2024comat,liang2024rich,wu2023human,xu2024imagereward,huang2023t2i,lin2025evaluating,gordon2025mismatch,lu2024llmscore,hu2023tifa,yarom2024you,chodavidsonian}.

\paragraph{Targeted corrections}
Many methods pursue \emph{targeted} fixes—identifying specific errors in a generated image and editing them (e.g., attribute repair or object adjustment) using editing-capable models~\cite{wu2024self,yang2024mastering}. Relatedly, several prompt-optimization approaches also rely on feedback from LLMs or MLLMs that explicitly surfaces which parts of the prompt fail to meet the desired alignment, driving the optimization through targeted correction at each step~\cite{wan2025maestro, khan2025test}. 

\paragraph{Multi-turn generation}
Several methods adopt iterative, multi-turn pipelines to better match user intent:
SLD~\cite{wu2024self} proposes coordinate and attribute edits, GenArtist~\cite{wang2024genartist} performs verification and self-correction with an MLLM, PASTA~\cite{nabati2024personalized} optimizes preferences via RL, and T2I-Copilot~\cite{chen2025t2i} uses an MLLM-based evaluation agent with optional human-in-the-loop control. NPC follows this multi-turn refinement paradigm by leveraging negative prompts to guide regeneration toward improved alignment.

\section{The Role of Negative Prompts}\label{sec:understanding}

In this section, we examine how negative prompting affects text–image alignment. We first describe its mechanism within T2I models through the CFG formulation (\Cref{sec3.1}), then introduce an attention-based score for quantifying its impact (\Cref{sec:attention_score}). Using this score, we show that both targeted and untargeted negative prompts contribute to improved alignment (\Cref{sec3.3}).

\subsection{Negative Prompting in T2I Models}\label{sec3.1}
Modern diffusion- and flow-based text-to-image models
\cite{rombach2022high,podellsdxl,esser2024scaling,flowmatching,rectifiedflow}
produce, at each timestep $t$, a prediction signal $f_\theta(x_t, c, t)$, conditioned on text $c$, a formulation that encompasses noise prediction (diffusion), velocity prediction (flow-matching), and score prediction within a single unified expression.

Let $p$ denote the \textit{positive} prompt and $n$ the \textit{negative} (or unconditional) prompt.
Classifier-free guidance (CFG) forms a guided estimate by contrasting the positive branch with a null-conditioned branch:
\begin{equation}
\label{eq:cfg-unified}
\hat{f}_t
=
f_\theta(x_t, n, t)
+
\lambda \Big(
        f_\theta(x_t, p, t)
        -
        f_\theta(x_t, n, t)
      \Big),
\end{equation}
where $\lambda$ is the guidance scale.
Replacing this null-conditioning with an explicit negative prompt preserves the classical CFG structure while enabling direct suppression of undesired concepts.
The latent is then updated through a generic sampler $\Phi_t$, yielding $x_{t-1} = \Phi_t(x_t, \hat{f}_t)$.

\begin{figure}[t]
\begin{center}
\includegraphics[width=\columnwidth]{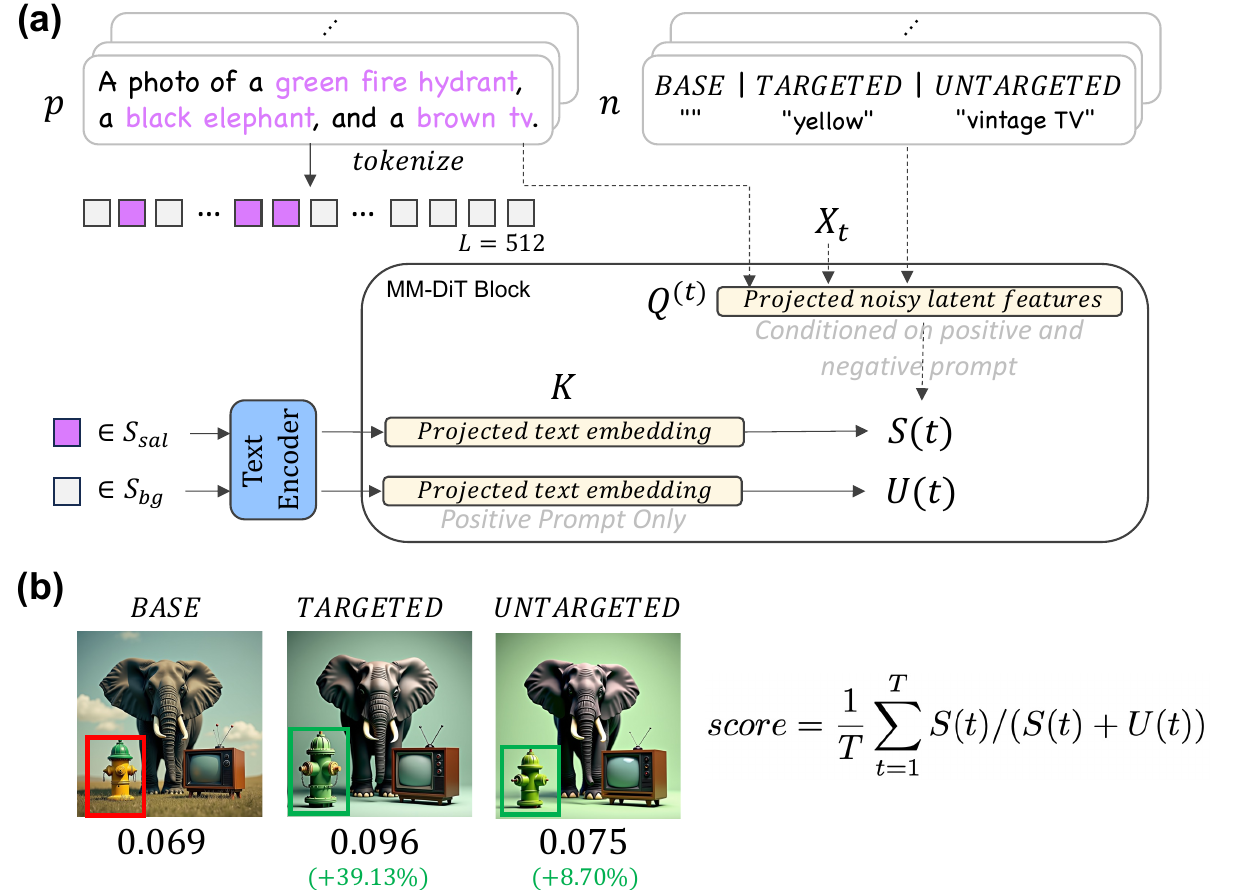}
  \end{center}
  \vskip -0.2in
  \caption{Analysis of how negative prompts influence alignment.
(a) shows that cross-attention is computed with positive-prompt keys and latents (queries) influenced by both prompts.
(b) shows that both negative prompts increase attention to salient tokens over BASE, illustrating their alignment-improving effect. Additional illustrative prompt examples are provided in~\Cref{sec:additional_result}.}\label{fig:sas}
\end{figure}
\begin{figure*}[t]
\begin{center}
\includegraphics[width=\textwidth]{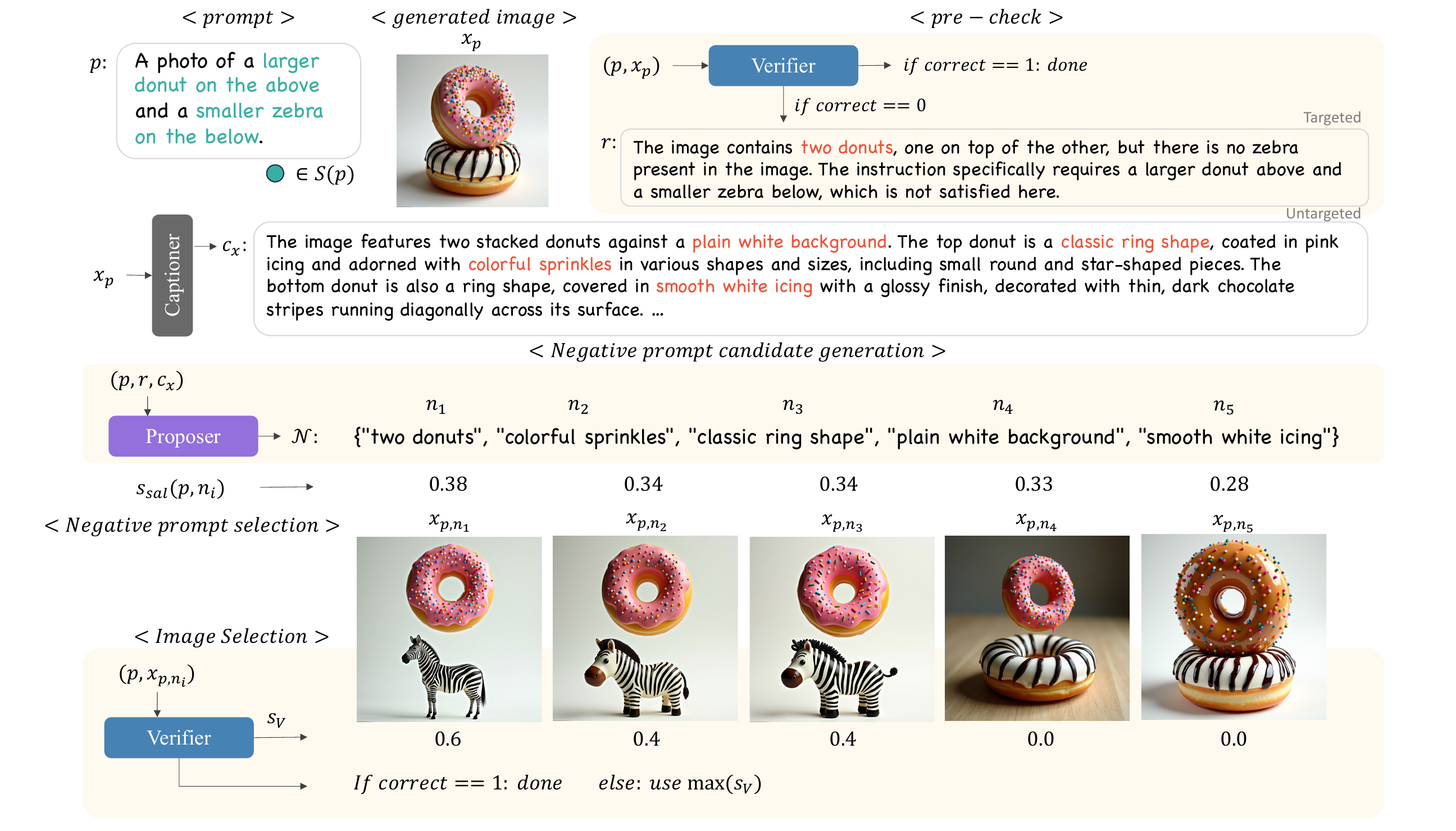}
  \end{center}
  \vskip -0.2in
  \caption{NPC overview. NPC runs an iterative generate–evaluate–refine loop: (i) a verifier judges each generated image and, if incorrect, produces a failure reason; (ii) a captioner generates a caption; (iii) the failure reason and caption condition a proposer to propose negative-prompt candidates; (iv) each candidate receives a salient score; candidates are applied in descending score order to regenerate images; verification is repeated, terminating upon a correct result or, if none are correct, selecting the sample with the highest verifier score.}\label{fig:npc}
\end{figure*}

\subsection{Measuring Cross-Attention Allocation}
\label{sec:attention_score}

Our objective is to quantify how negative prompting alters the allocation of attention toward the semantics of the positive prompt. To this end, we separate the roles of queries and keys in the text--image cross-attention layers of Transformer-based backbones used in diffusion and flow-matching models (e.g., DiT architectures~\cite{labs2025flux,dit}).

\paragraph{Cross-attention}
After applying the guided update in~Equation~\eqref{eq:cfg-unified}, we extract
image-stream hidden states $H^{(t)} = \mathrm{ImgEnc}(x_{t-1})$, which serve as
the cross-attention queries and encode the influence of both the positive and
negative prompts. To keep the text-side conditioning fixed, we obtain token
embeddings of the positive prompt $E(p) \in \mathbb{R}^{L \times d_e}$ and form
keys as $K^{\mathrm{cond}} = W_k E(p)$. The queries, in contrast, evolve with
the latent trajectory through $Q^{(t)} = W_q H^{(t)}$.
For batch index $b$, head $h$, spatial position $q$, and token index $k$, the attention logit is
\begin{equation}
    s^{(t)}_{b,h,q,k}
    =
    \frac{
        Q^{(t)}_{b,h,q} \cdot K^{\mathrm{cond}}_{b,h,k}
    }{
        \sqrt{d_h}
    }.
\end{equation}
Softmax over tokens yields the attention probabilities:
\begin{equation}
    P^{(t)}_{b,h,q,k}
    =
    \frac{\exp(s^{(t)}_{b,h,q,k})}{\sum_{j}\exp(s^{(t)}_{b,h,q,j})}.
\end{equation}

\paragraph{Salient-attention score}
Let $\mathcal{S}_{\mathrm{sal}} \subset \{1,\dots,L\}$ be the set of salient
tokens in the positive prompt—those corresponding to semantic elements that
directly determine alignment quality (e.g., objects or their attributes). The
remaining tokens form $\mathcal{S}_{\mathrm{bg}}$. Since proper alignment
requires the model to allocate sufficient attention to these salient tokens, an
increase in their attention share reflects improved semantic grounding. We aggregate cross-attention mass over batch, heads, and spatial positions:
\[
S(t)=\sum_{b,h,q}\sum_{k\in\mathcal{S}_{\mathrm{sal}}}P^{(t)}_{b,h,q,k},
\qquad
U(t)=\sum_{b,h,q}\sum_{k\in\mathcal{S}_{\mathrm{bg}}}P^{(t)}_{b,h,q,k}.
\]
The \textit{salient-attention score} is then defined as
\begin{equation} \rho_{\mathrm{sal}} = \frac{1}{T} \sum_{t=1}^{T} \frac{S(t)}{S(t)+U(t)}  \end{equation}
where $T$ denotes the total number of denoising steps.
This score reflects how much of the model’s cross-attention is allocated to alignment-critical tokens. Since keys come from the positive prompt and only queries change under negative prompting, variations in $\rho_{\mathrm{sal}}$ reveal how negative prompts shift the model’s semantic focus throughout the denoising process.

\subsection{Analysis Result}\label{sec3.3}

For the prompt “A photo of a \emph{green} fire hydrant, a black elephant, and a brown tv,” we first generate an image using the BASE configuration, which applies no negative prompt. As shown in~\Cref{fig:sas}(b), the hydrant appears \emph{yellow} rather than green, revealing an attribute error. In this setting, the incorrect attribute naturally yields a targeted negative prompt (“yellow fire hydrant”), while an untargeted negative is chosen from incidental attributes present in the image but irrelevant to the error (e.g., “vintage television”).

The BASE model attains a salient-attention score of 0.069, which increases to 0.096 with targeted negative guidance (39.13\%) and to 0.075 with an untargeted negative (8.70\%). Rather than merely correcting the hydrant’s appearance, these changes reflect how negative prompting reshapes the model’s internal focus: the targeted negative suppresses the erroneous attribute, enabling attention to shift back toward the intended “green,” while the untargeted negative reduces emphasis on unrelated background elements. Together, these results show that negative prompts—whether directly or indirectly related to the misalignment—can reallocate attention toward the prompt's salient components and thereby improve alignment. Although both types of negatives help, their effectiveness varies across prompt–image pairs $(p,x)$, making manual selection difficult and motivating an automated mechanism.

\section{NPC}
In this section, we introduce NPC, an automated pipeline for selecting effective negative prompts by (i) generating candidates (\Cref{sec:candidate_generation}) and (ii) scoring them with a salient text-space measure that requires no additional image generation (\Cref{sec:selection}). \Cref{fig:npc} provides an overview.



\subsection{Negative Prompt Candidate Generation}\label{sec:candidate_generation}
\begin{figure}[t]
  \begin{center}
\includegraphics[width=\columnwidth]{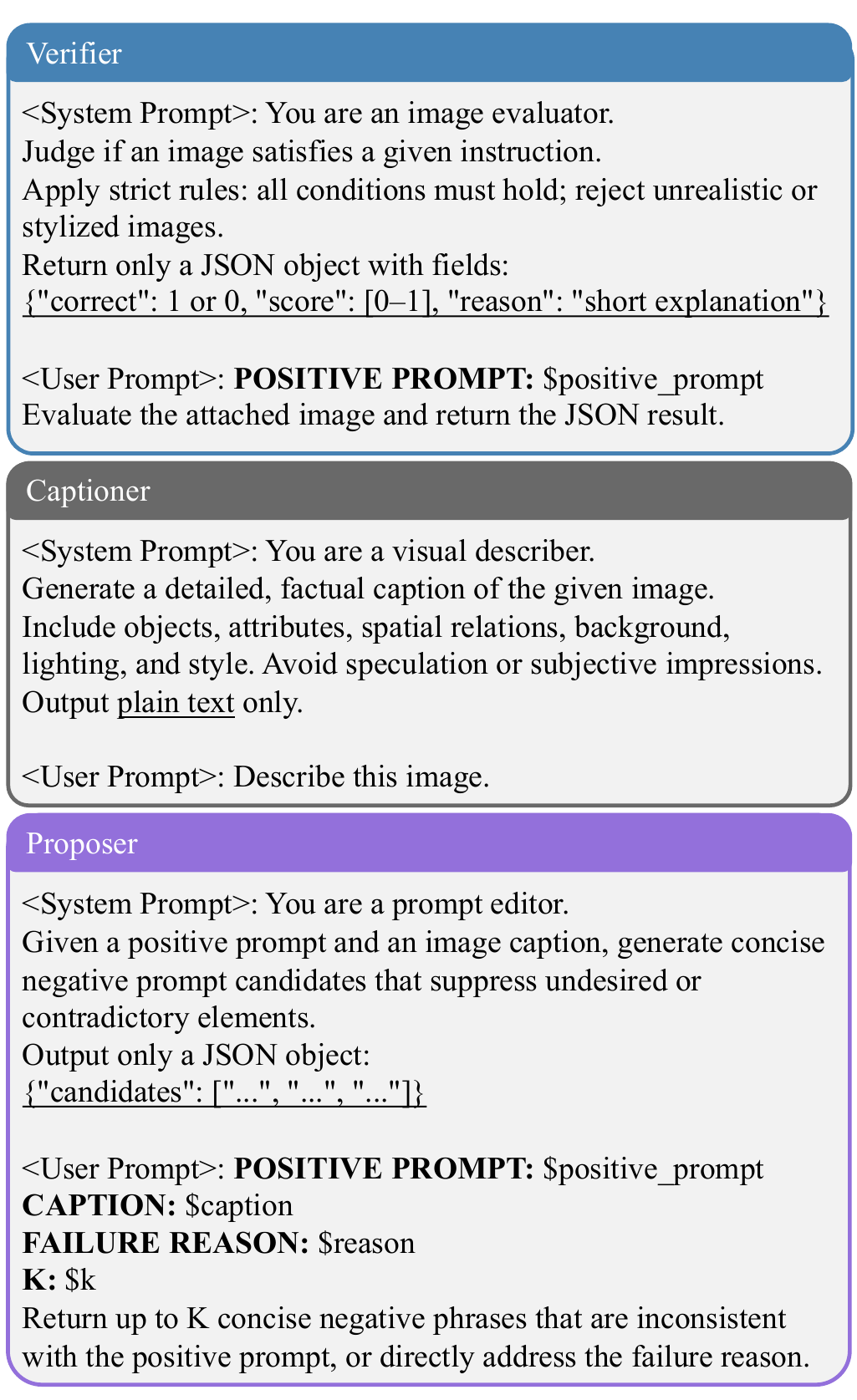}
  \end{center}
  \vskip -0.15in
   \caption{Summarized LLM prompts for each agent (see \Cref{sec:llm_prompt} for full versions). Inputs are shown in bold in the user prompt, while outputs are underlined in the system prompt.}
   \label{fig:llm_prompt}
   
\end{figure}

As the efficacy of negative prompts is unknown \emph{a priori}, it is natural to generate a set of candidates. To cover the full design space—including both \emph{targeted} and \emph{untargeted} negatives—we employ distinct agents with complementary roles. Given an positive prompt \(p\) and its generated image \(x_p\), \textsc{NPC} uses three agents—the \emph{verifier}, \emph{captioner}, and \emph{proposer}—to produce negative–prompt candidates. A concise version of the LLM prompts that define each agent’s role is shown in \Cref{fig:llm_prompt}.

\begin{itemize}
    \item \textbf{Verifier} \(V\) checks whether \(x_p\) satisfies alignment criteria w.r.t.\ \(p\) and returns \((\texttt{correct}, s_V, r)\), where \(\texttt{correct}\in\{0,1\}\), \(s_V\in[0,1]\) is a confidence score, and \(r\) is a short \texttt{reason} string (e.g., missing attributes, wrong style, prohibited content) that can serve as \emph{targeted} negatives.
    \item \textbf{Captioner} \(C\) produces a descriptive caption \(c_x\) of \(x_p\), enumerating incidental objects, attributes, and context to supply \emph{untargeted} negatives.
    \item \textbf{Proposer} \(P\) distills \(r\) and \(c_x\) into concise (typically one– or two–token) negative prompts. Conditioning on \((p,r,c_x)\), it proposes \(K\in\mathbb{N}\) candidates \(\mathcal{N}=\{n_1,\dots,n_K\}\). Each \(n_k\) is either \emph{targeted}—directly addressing \(r\)—or \emph{untargeted}—suppressing incidental scene elements hypothesized to interfere with alignment.
\end{itemize}

\paragraph{Pipeline}
We first generate \(x_p\) and query \(V(x_p,p)\).
If \(\texttt{correct}=1\), we skip \textsc{NPC} and return \(x_p\) as is (pre-check in \Cref{fig:npc}).
Otherwise, we obtain \(r\gets V(x_p,p)\) and \(c_x\gets C(x_p)\), then form \(\mathcal{N}\gets P(p,r,c_x)\).
\Cref{sec:selection} details \emph{automated effective negative prompt selection}, in which candidates in \(\mathcal{N}\) are \emph{scored in text space} (without additional image synthesis) and the top-scoring prompt is selected.

\begin{table*}[t]
\centering
\setlength{\tabcolsep}{8pt}
\renewcommand{\arraystretch}{1.2}
\resizebox{\textwidth}{!}{%
\begin{tabular}{lcccccccc}
\toprule
\textbf{Method} & \textbf{Color} & \textbf{Count} & \textbf{Color/Count} & \textbf{Color/Pos} & \textbf{Pos/Count} & \textbf{Pos/Size} & \textbf{Multi-Count} & \textbf{Overall}\\
\midrule
SDv2.1 {\cite{rombach2022high}}        & 0.000 & 0.325 & 0.025 & 0.000 & 0.000 & 0.025 & 0.075 & 0.064 \\
SDXL {\cite{podellsdxl}}               & 0.050 & 0.375 & 0.000 & 0.000 & 0.000 & 0.000 & 0.000 & 0.061 \\
SDv3-medium {\cite{esser2024scaling}}  & \underline{0.550} & 0.500 & 0.125 & 0.350 & 0.175 & 0.150 & 0.225 & 0.296 \\
FLUX.1-Kontext {\cite{labs2025flux}}   & 0.425 & 0.500 & 0.200 & 0.250 & \underline{0.300} & 0.400 & 0.325 & 0.343 \\
FLUX.1-dev {\cite{black-forest-labs-no-date}} & 0.350 & \underline{0.625} & 0.150 & 0.275 & 0.200 & 0.375 & 0.225 & 0.314 \\
\midrule
Janus-Pro {\cite{chen2025janus}}       & 0.450 & 0.300 & 0.125 & 0.300 & 0.075 & 0.350 & 0.125 & 0.246 \\
T2I-R1 {\cite{jiang2025t2i}}           & \textbf{0.675} & 0.325 & 0.200 & 0.350 & 0.075 & 0.250 & 0.300 & 0.311 \\
BLIP3-o 4B {\cite{chen2025blip3}}      & 0.125 & 0.225 & 0.100 & 0.450 & 0.125 & 0.550 & 0.225 & 0.257 \\
BLIP3-o 8B {\cite{chen2025blip3}}      & 0.250 & 0.250 & 0.125 & \textbf{0.600} & 0.125 & \underline{0.575} & 0.225 & 0.307 \\
OmniGen2 {\cite{wu2025omnigen2}}       & \underline{0.550} & 0.425 & 0.200 & 0.275 & 0.125 & 0.250 & \underline{0.450} & 0.325 \\
Bagel {\cite{deng2025emerging}}        & 0.325 & 0.600 & \underline{0.250} & 0.325 & 0.250 & 0.475 & 0.375 & \underline{0.371} \\
\rowcolor{blue!10}
\textbf{NPC (Ours)} & \underline{0.550} & \textbf{0.675} & \textbf{0.350} & \underline{0.525} & \textbf{0.550} & \textbf{0.725} & \textbf{0.625} & \textbf{0.571} \\
\bottomrule
\end{tabular}
}%
\caption{Comparative results on GenEval++ assessing diverse compositional generation capability. Each sample is scored 0/1 based on correctness, so the reported values are accuracies. Bold indicates the best result, and underlined denotes the second best (ties underlined).}
\label{tab:GenEval++}
\end{table*}

\begin{table*}[t]
\centering
\renewcommand{\arraystretch}{1.2}
\footnotesize
\begin{tabular}{lccccc}
\toprule
\textbf{Method} & \textbf{Attribute shift} & \textbf{Spatiotemporal} & \textbf{Hybridization} & \textbf{Multi-Object} & \textbf{Overall} \\
\midrule
SDv2.1~\cite{rombach2022high}          & 4.46 & 5.06 & 4.12 & 3.49 & 4.30 \\
SDXL~\cite{podellsdxl}                  & 4.42 & 6.32 & 4.93 & 4.50 & 4.97 \\
SDv3-medium~\cite{esser2024scaling}     & 5.14 & 5.91 & 6.30 & 6.07 & 5.78 \\
FLUX.1-Kontext~\cite{labs2025flux}      & 5.33 & 6.49 & 5.48 & 5.34 & 5.62 \\
FLUX.1-dev~\cite{black-forest-labs-no-date} & 5.68 & 7.13 & 6.38 & 5.24 & 6.06 \\
\midrule
Janus-Pro~\cite{chen2025janus}          & 5.30 & 7.28 & 6.73 & 6.04 & 6.22 \\
T2I-R1~\cite{jiang2025t2i}              & \underline{5.85} & \textbf{7.70} & \underline{7.36} & \textbf{6.68} & \underline{6.78} \\
BLIP3-o 4B~\cite{chen2025blip3}         & 5.48 & 6.79 & 6.93 & 6.09 & 6.23 \\
BLIP3-o 8B~\cite{chen2025blip3}         & 5.80 & 7.08 & 7.06 & 6.44 & 6.51 \\
OmniGen2~\cite{wu2025omnigen2}          & 5.28 & \underline{7.45} & 6.29 & 6.31 & 6.22 \\
Bagel~\cite{deng2025emerging}           & 5.37 & 6.93 & 6.50 & 6.41 & 6.20 \\
\rowcolor{blue!10}
\textbf{NPC (Ours)}                     & \textbf{6.31} & 6.76 & \textbf{7.57} & \underline{6.52} & \textbf{6.80} \\
\bottomrule
\end{tabular}
\caption{Comparative results on Imagine-Bench assessing the ability to generate surreal modifications. A detailed description of the scoring is provided in \Cref{sec:experimental_setting}. Bold indicates the best result; underlined denotes the second best.}
\label{tab:Imagine}
\end{table*}

\subsection{Negative Prompt Selection}\label{sec:selection}
To minimize re-generation trials without incurring additional compute or latency, \textsc{NPC} introduces a proxy \emph{text-space} scorer—the \emph{salient score}—which estimates how a negative prompt \(n\) will affect alignment with the salient content of the positive prompt \(p\) using only the model’s text encoder.  
Intuitively, the score (i) models how adding \(n\) shifts the text embedding of \(p\), (ii) identifies salient tokens in \(p\) (e.g., key objects or attributes), and (iii) measures how well this shift aligns with those salient tokens via cosine similarity.  
Averaging these similarities yields a scalar score that ranks negative candidates without requiring any image synthesis.  
Although this text-space measure is not an exact proxy for image-level effects, it is effective for comparative evaluation and candidate selection; see \Cref{sec:ablation_studies} for empirical evidence.

Formally, let \(\mathbf{E}(\cdot)\in\mathbb{R}^{L\times d}\) denote token-level outputs of the text encoder~\cite{raffel2020exploring}, and let \(\mathrm{pool}(\cdot)\) be a fixed pooling operator (mean over tokens).  
We form pooled embeddings \(\bar{\mathbf e}_p=\mathrm{pool}(\mathbf{E}(p))\) and \(\bar{\mathbf e}_n=\mathrm{pool}(\mathbf{E}(n))\), and use cosine similarity  $\cos(\mathbf a,\mathbf b)=\langle \mathbf a,\mathbf b\rangle/(\|\mathbf a\|\;\|\mathbf b\|).$
The \emph{direction of change} induced by adding \(n\) is modeled as the subtractive direction
$\mathbf d(p,n)=\bar{\mathbf e}_p-\bar{\mathbf e}_n,$
treating \(n\) as pulling the representation away from content associated with \(n\).

We extract \emph{salient tokens} from a prompt using a simple heuristic: we identify \emph{quantity tokens} (e.g., ``a'', ``two'') and take the noun that immediately follows each quantity token as a salient-token candidate.  
For example, for the prompt  
``A jelly-like dolphin and a steel-reinforced dam coexist in the gentle riverbed flow,''  
our procedure yields $\mathcal{S}_{\mathrm{sal}}(p) = \{\text{jelly-like},\; \text{steel-reinforced}\}.$
For each \(t\in\mathcal{S}_{\mathrm{sal}}(p)\), define \(\mathbf v_t=\mathrm{pool}(\mathbf{E}(t))\) and \(s_t=\cos(\mathbf d(p,n),\mathbf v_t)\).  
The \emph{salient score} is then
\[
s_{\mathrm{sal}}(p,n)
\triangleq
\frac{1}{|\mathcal{S}_{\mathrm{sal}}(p)|}\sum_{t\in \mathcal{S}_{\mathrm{sal}}(p)} s_t \, .
\]
A higher \(s_{\mathrm{sal}}(p,n)\) indicates that \(\mathbf d(p,n)\) is better aligned with the salient content of \(p\)—i.e., the negative tends to preserve core semantics and is more likely to be alignment-preserving. Accordingly, we process candidates in decreasing order of \(s_{\mathrm{sal}}(p,n)\) and synthesize images following this order (with early stopping once alignment is satisfied).

\paragraph{Stopping rule}
Given candidates \(\{n_i\}_{i=1}^K\) (in the prescribed order), for \(i=1,\dots,K\) generate \(x_{p, n_i}\) under \((p,n_i)\) and query \(V(x_{p, n_i},p)\).
Terminate at the first \(i\) with \(\texttt{correct}=1\) and output \(x_i\).
If no such \(i\) exists, fallback to the image with the highest verifier score $s_{V_i}$, i.e., output \(x_{p, n_i^\star}\) where \(i^\star=\arg\max_{i}{s_{V_i}}\).

\begin{figure*}[t]
\begin{center}
\includegraphics[width=0.9\textwidth]{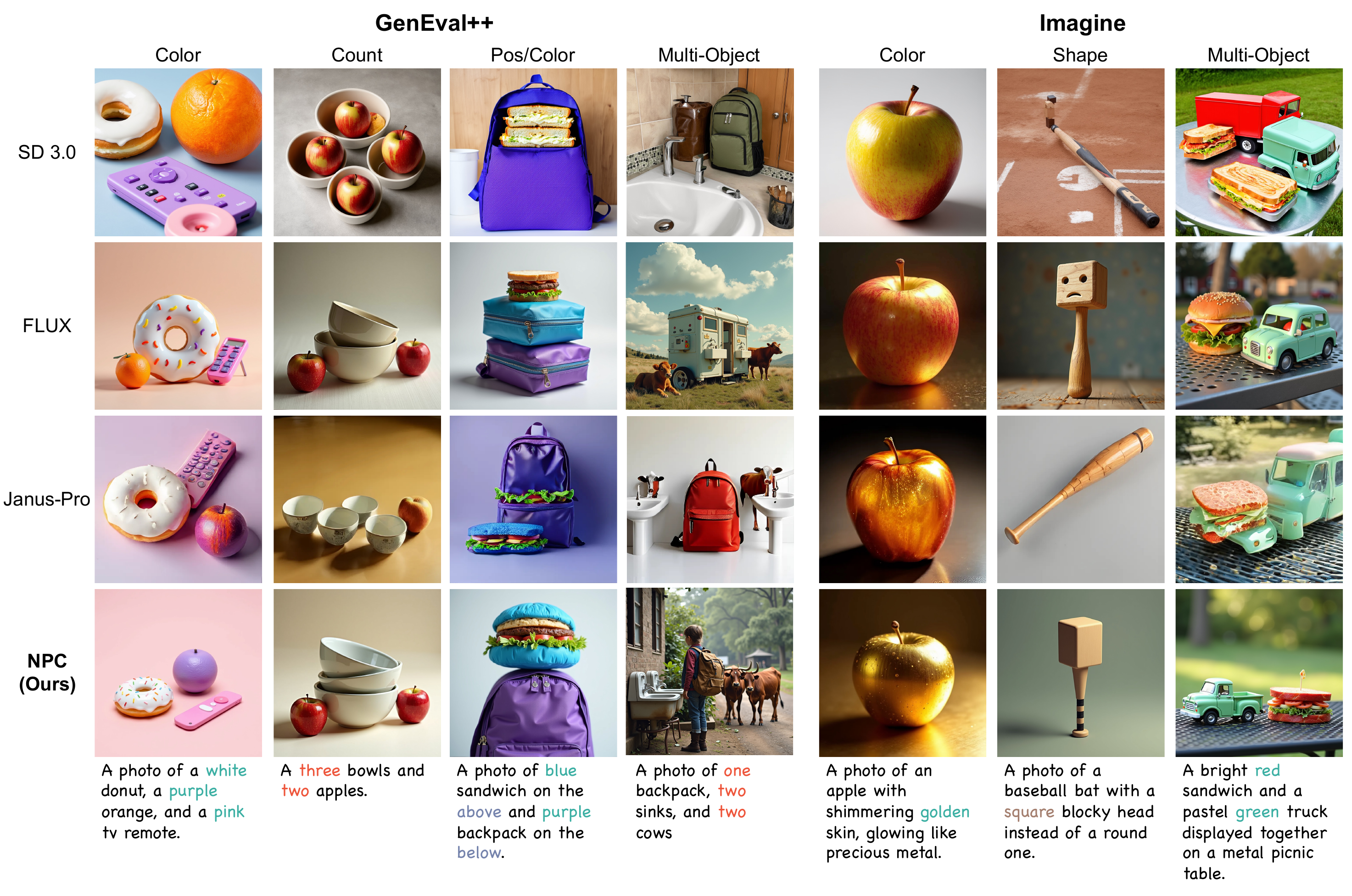}
  \end{center}
  \vskip -0.2in
  \caption{Qualitative comparison on GenEval++ and Imagine-Bench: NPC (samples generated with FLUX) consistently achieves alignment across diverse tasks, while other methods fail in some cases.}\label{fig:baselines}
\end{figure*}
\section{Experiments}
\subsection{Experimental Setting}\label{sec:experimental_setting}
\paragraph{Implementation details} 
We instantiate the image generator with \textsc{FLUX}.1-dev~\cite{black-forest-labs-no-date}. The \emph{Captioner} and the \emph{Proposer} are implemented with GPT-4o-mini~\cite{openai2025gpt4o}, and the \emph{Verifier} uses GPT-4.1~\cite{openai2025gpt41}. Additional experiments using other image generators and alternative MLLM or LLM configurations can be found in~\Cref{sec:model_agnostic}. For each prompt \(p\), the \emph{Proposer} returns \(K=5\) candidates \(\mathcal N=\{n_1,\dots,n_5\}\). Diffusion sampling uses \(T=50\) denoising steps; to reduce computation, we apply the negative prompt only in the first three (\(t=1,2,3\)), as later steps are empirically observed to have negligible influence. Unless otherwise noted, we set the guidance scale to 3.5 and the true-CFG scale (a variant of classifier-free guidance that uses the direct difference between conditional and unconditional prediction) to 1.8.

\paragraph{Datasets and evaluation}
We evaluate two benchmarks, recently proposed by~\cite{ye2025echo}, that assess instruction fidelity and creative synthesis.

\textbf{GenEval++} This benchmark provides 280 prompts (seven task types with 40 prompts each) featuring richer semantics and more diverse compositions than prior instruction–following sets. A GPT\mbox{-}4.1~\cite{openai2025gpt41} multimodal evaluator assesses image–text consistency using a fixed checklist over \emph{Object}, \emph{Counts}, \emph{Color}, \emph{Position}, and \emph{Size}; a result is marked correct only if all criteria are satisfied. To match the “A photo of …” prompt style, anime–style renderings and images with multiple disjoint elements are deemed invalid.

\textbf{Imagine-Bench} This benchmark contains 270 creative instructions that modify common objects with fantastical attributes while preserving identity. For each instruction, GPT\mbox{-}4o~\cite{openai2025gpt4o} produces a checklist specifying the required fantastical modifications and the object’s invariant identity cues. Given the prompt and image, GPT\mbox{-}4.1 assigns 0–10 scores on three dimensions—\emph{Fantasy Fulfillment}, \emph{Identity Preservation}, and \emph{Aesthetic Quality}—with explicit rationales. The final score is computed as $0.8 \times \min(\text{Fantasy Fulfillment},\, \text{Identity Preservation}) \;+\; 0.2 \times \text{Aesthetic Quality}$.
Additional text-image alignment benchmark results—DPG-Bench~\cite{hu2024ella}, and T2I-CompBench~\cite{huang2023t2i}—are in \Cref{sec:additional_benchmark}.

\paragraph{Baselines}
We compare NPC against four groups, all of which are strong baselines: (i) \textbf{Stable Diffusion series}—SD~v2.1~\cite{rombach2022high}, SDXL~\cite{podellsdxl}, and SD3-Medium~\cite{esser2024scaling}—strong open T2I diffusion baselines; (ii) \textbf{FLUX.1 family}—FLUX.1-dev~\cite{black-forest-labs-no-date} and FLUX.1-Kontext~\cite{labs2025flux}—rectified-flow Transformer T2I models, with \emph{Kontext} enabling in-context conditioning/editing; (iii) \textbf{Unified multimodal generators}—BLIP3-o (4B/8B)~\cite{chen2025blip3}, OmniGen2~\cite{wu2025omnigen2}, and Bagel~\cite{deng2025emerging}—models trained for joint understanding and generation via large-scale pretraining; and (iv) \textbf{Reasoning/AR frameworks}—Janus-Pro~\cite{chen2025janus} and T2I-R1~\cite{jiang2025t2i}—autoregressive or unified approaches that incorporate chain-of-thought or RL signals to enhance instruction following. 
Additional comparison results against other text–image alignment baselines are provided in~\Cref{sec:additional_benchmark}.

\begin{figure}[t]
  \begin{center}
\includegraphics[width=\columnwidth]{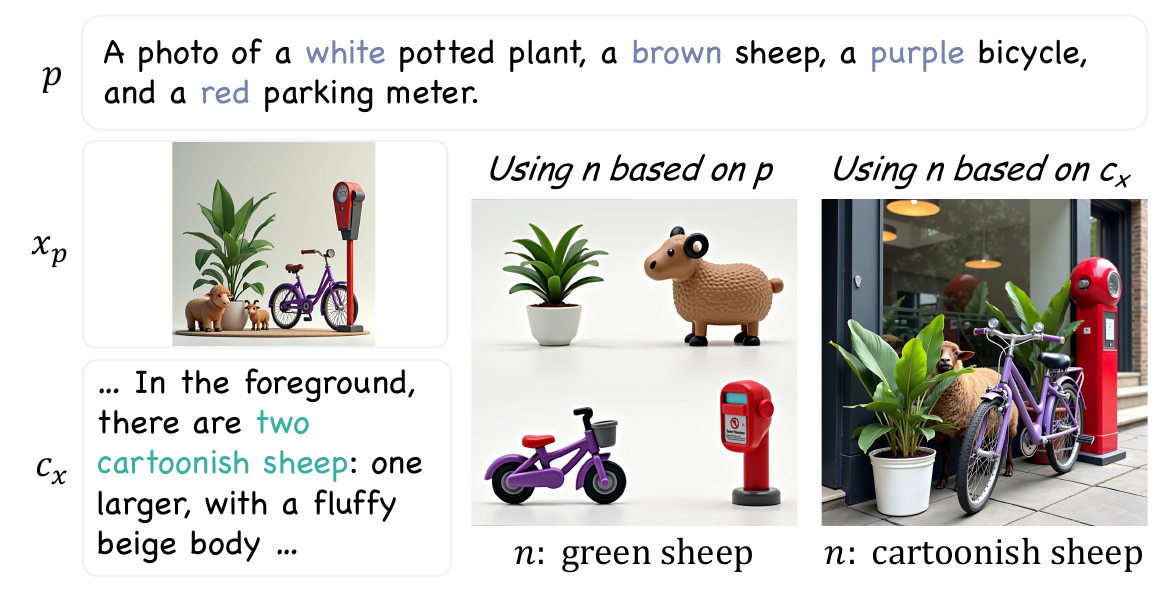}
  \end{center}
  \vskip -0.2in
   \caption{Ablation of negative–prompt generation (\Cref{sec:ablation_studies}): negatives generated from the original prompt ($p$) vs. those generated from the original prompt with the caption additionally provided (NPC). NPC yields the more realistic result.}
   \label{fig:prompt_ablation}
\end{figure}

\subsection{Main Results}
\Cref{tab:GenEval++} and \Cref{tab:Imagine} collectively demonstrate that NPC attains state-of-the-art performance across both benchmarks. On GenEval++ (Table~\ref{tab:GenEval++}), NPC achieves the highest overall score (0.571), surpassing the strongest unified baseline, Bagel (0.371), with pronounced gains on positional/compositional constraints—Count/Pos, Pos/Size, and Multi-Count. For Color and Color/Pos, NPC ranks second best—narrowly trailing the strongest baselines—reflecting competitive, near-leading performance in these dimensions. On Imagine-Bench (Table~\ref{tab:Imagine}), NPC achieves the best overall score (6.80) and ranks first in all categories except \emph{Spatiotemporal}, where T2I-R1 leads. Overall, these results indicate that NPC performs robustly across diverse task types on Imagine-Bench; qualitative comparisons with recent T2I models are shown in~\Cref{fig:baselines}.

\subsection{Ablation Study}\label{sec:ablation_studies}
This section presents ablations of the two core components of NPC: (i) negative–prompt candidate generation and (ii) selection based on a salient text–space metric. We evaluate each component on GenEval++ and quantify its impact on alignment quality and computational efficiency.

\paragraph{Negative Prompt Generation}
To incorporate information from the original image, we condition the \emph{proposer} on a \textbf{caption} when generating negatives. For validation, we compare NPC (caption-conditioned) with a prompt-only baseline that proposes negatives from the original prompt using a separate \emph{proposer} (LLM instructions in \Cref{tab:prompt_ablation}). NPC marginally but consistently outperforms the prompt-only baseline on GenEval++—0.57 vs.\ 0.53 accuracy—indicating that grounding negatives in actual image content is more effective. This ablation is designed to assess the utility of the \emph{captioner}; consistent with the quantitative result, \Cref{fig:prompt_ablation} shows that grounded negatives (e.g., “cartoonish sheep”) produce more realistic outputs than negatives targeting attributes not supported by the image (e.g., “green sheep”).

\paragraph{Salient Score}

\begin{figure}[t]
  \begin{center}
\includegraphics[width=\columnwidth]{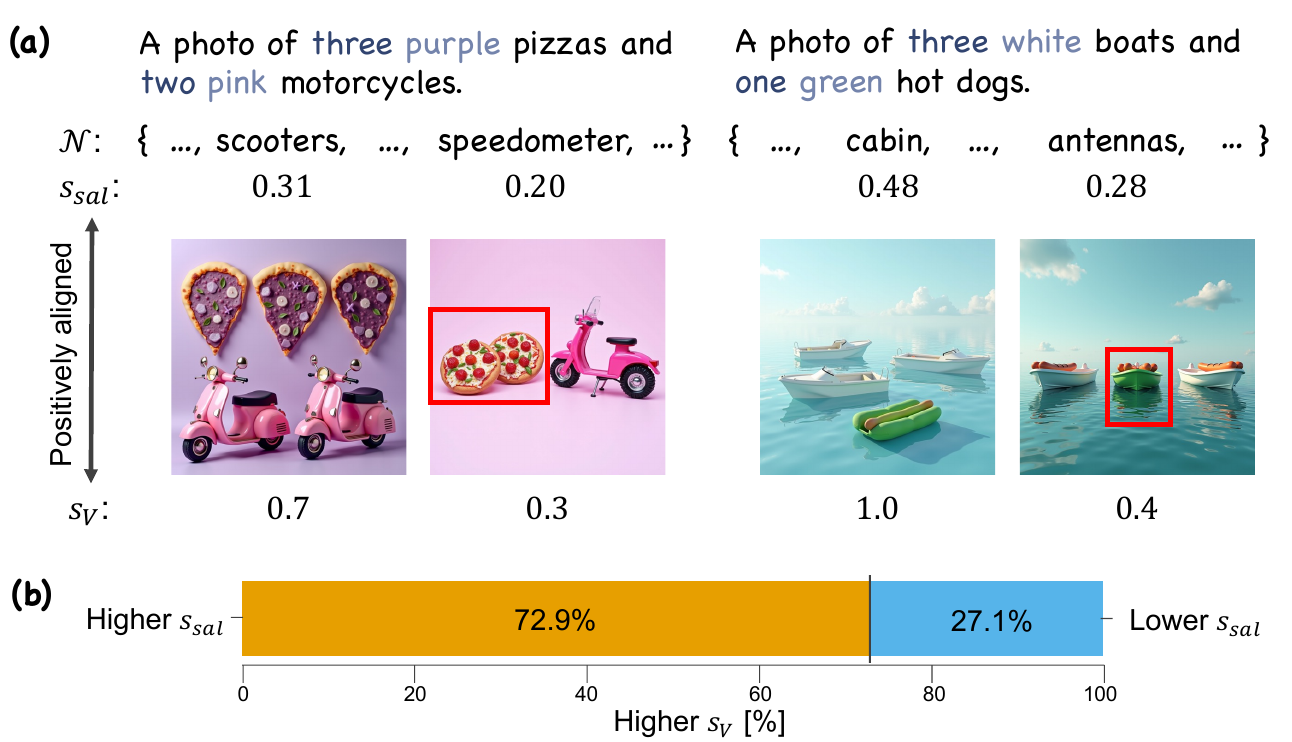}
  \end{center}
  \vskip -0.1in
   \caption{Examples of relationship between salient score ($s_{sal}$) and verifier score ($s_V$) (\Cref{sec:ablation_studies}); the higher-salient negative yields the higher verifier score.}
   \label{fig:rank}
\end{figure}

To assess how useful the salient score is as a priority signal for the selection step of NPC, we run two tests: (1) \textbf{regeneration efficiency}—holding the candidate set fixed (with five candidates), we compare descending salient-score order against a random order; random ordering can push an effective negative to the back, increasing total generations, whereas prioritization lowers the mean number of attempts from \(4.1\) to \(2.5\); and (2) \textbf{alignment quality}—for each prompt, we choose two negatives with different priorities, generate one image per negative under identical conditions, and compare verifier scores; the higher-salience negative attains the higher verifier score in \(204/280\) pairs (\(72.9\%\); examples in \Cref{fig:rank}). These results indicate that the salient score provides an effective and practical priority signal for negative prompt selection, both reducing regeneration and improving alignment quality.

\section{Conclusion}
We propose \textbf{NPC}, an automated pipeline that improves text--to--image alignment by discovering and selecting \emph{effective negative prompts}. Beyond resolving explicit prompt conflicts, our image--text attention analysis shows that \emph{untargeted} negatives---derived from incidental visual attributes---also strengthen alignment. NPC consists of (i) a verifier--captioner--proposer framework that generates both targeted and untargeted negatives, and (ii) a selection stage that ranks candidates with a salient text-space metric to minimize regeneration. In experiments, NPC yields strong alignment gains with modest generation overhead. These results highlight the central role of negative-prompt design in accurate, reliable image generation and point to promising directions for further progress.
\section*{Acknowledgements}
This work was supported by the National Research Foundation of Korea (NRF) grant funded by the Korea government (MSIT) [No.2022R1A3B1077720; No.2022R1A5A7083908], Institute of Information \& Communications Technology Planning \& Evaluation (IITP) grant funded by the Korea government (MSIT) [No.RS-2025-02263754; No.RS-2022-II220959; No.RS-2021-II211343, Artificial Intelligence Graduate School Program (Seoul National University)], the BK21 FOUR program of the Education and Research Program for Future ICT Pioneers, Seoul National University in 2025. This research was also conducted as part of the Sovereign AI Foundation Model Project (Data Track), organized by MSIT and supported by the National Information Society Agency (NIA) of Korea [No.2025-AI Data-wi43].
{
    \small
    \bibliographystyle{ieeenat_fullname}

}
\clearpage

\appendix
\clearpage
\setcounter{page}{1}
\setcounter{section}{0}
\renewcommand\thesection{\Alph{section}}
\setcounter{table}{0}
\renewcommand{\thetable}{S\arabic{table}}
\setcounter{figure}{0}
\renewcommand{\thefigure}{S\arabic{figure}}
\maketitlesupplementary
\section{Additional Benchmark}\label{sec:additional_benchmark}
\paragraph{T2I-CompBench} 
T2I-CompBench~\cite{huang2023t2i} provides compositional prompts with multiple objects. We evaluate six tasks—attribute binding (color, shape, texture), object relationships (spatial, non-spatial), and a complex setting combining them—using 300 images per task (1,800 total). Metrics follow the benchmark: BLIP-VQA~\cite{huang2023t2i} for attribute binding, UniDet~\cite{zhou2022simple} for spatial relations, and CLIPScore~\cite{radford2021learning, hessel2021clipscore} for non-spatial relations; all three are applied to the complex task. In this setup, we compare our approach against recent alignment-enhancing methods in three groups: (1) \textit{attention map–based} (Attend-and-Excite~\cite{chefer2023attend}, SynGen~\cite{rassin2024linguistic}); (2) \textit{layout-based} (LMD~\cite{lian2024llmgrounded}, InstanceDiffusion~\cite{wang2024instancediffusion}); and (3) \textit{feedback-based} (GORS~\cite{huang2023t2i}, CoMat~\cite{jiang2024comat}). Only Attend-and-Excite is implemented on Stable Diffusion 2; the others (and ours) are implemented on SDXL. 

\begin{figure*}[t]
\begin{center}
\includegraphics[width=0.9\textwidth]{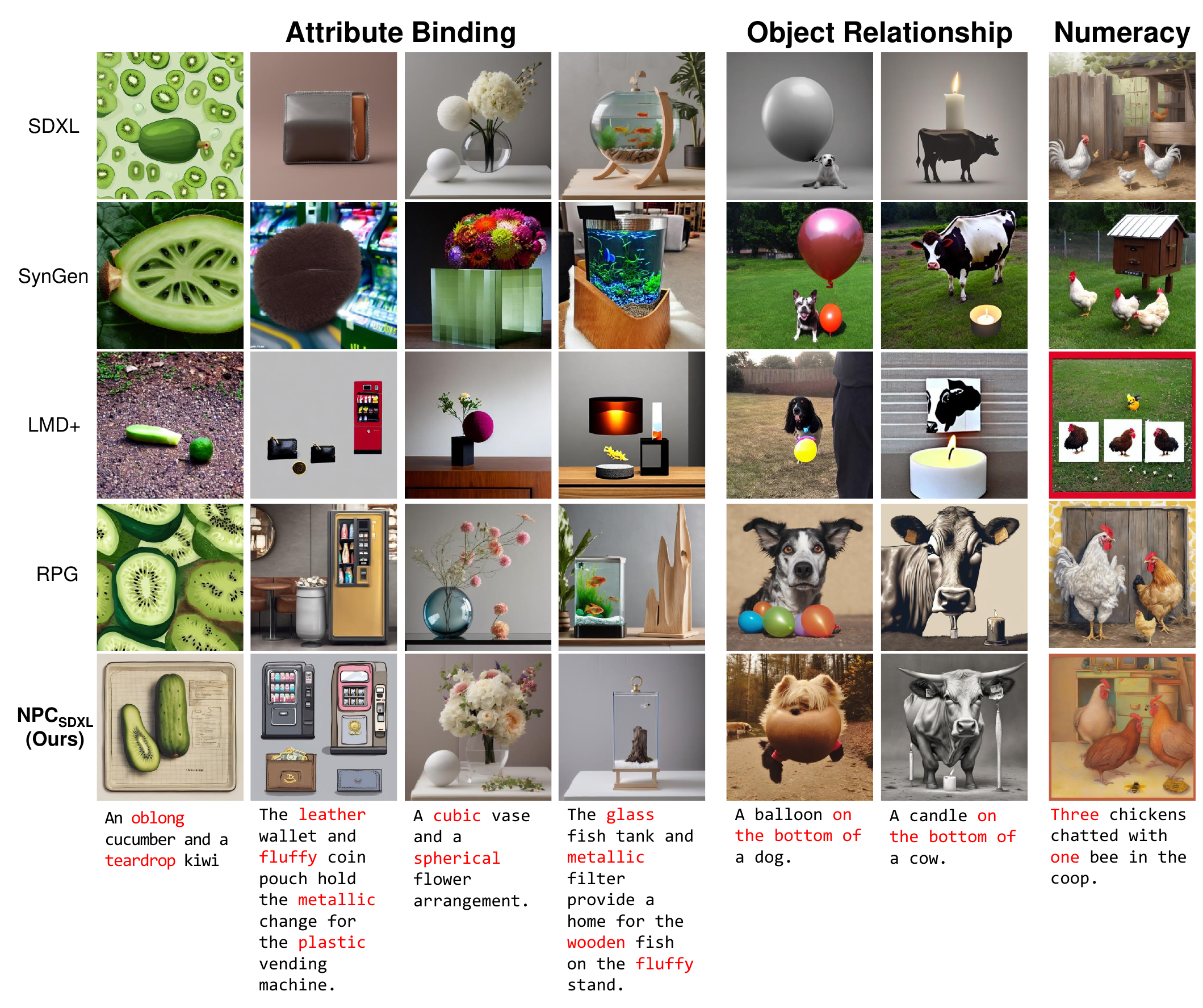}
  \end{center}
  \vskip -0.2in
  \caption{Qualitative comparison with SDXL~\cite{podellsdxl} and other text-image alignment methods~\cite{
  rassin2024linguistic, lian2024llmgrounded, yang2024mastering}. NPC consistently aligns across various tasks, while other methods fail in some tasks. Samples for NPC are generated using SDXL.}\label{fig:t2i_comp}
\end{figure*}

In this experiment, we implement NPC on SDXL for a fair comparison. As shown in \Cref{tab:maintable}, NPC attains the best attribute-binding performance and nearly a 15\%p gain on the complex task. While layout-based methods improve spatial controllability, their dependence on LLM outputs leads to failures with small or overlapping boxes~\cite{park2025raretofrequent}. In contrast, NPC avoids these issues, requires no additional training (unlike feedback-based methods), and needs no gradient updates (unlike attention-based methods), yielding both efficiency and superior average performance across tasks. Qualitative samples of alignment methods are also provided in~\Cref{fig:t2i_comp}.

\paragraph{DPG-Bench} 
DPG-Bench~\cite{hu2024ella} comprises 1,065 long, dense prompts designed for compositional text–image evaluation, drawn from COCO, PartiPrompts, DSG-1k, and Objects365 and expanded via GPT-4, then human-verified (5 level-1 and 13 level-2 categories).
DPG-Bench assesses text-to-image models on complex
prompts, using mPLUG-large~\cite{li-etal-2022-mplug} to
evaluate entities, attributes, and relationships with
a focus on fine-grained details like object properties and spatial arrangements.

\begin{figure*}[t]
\begin{center}
\includegraphics[width=0.9\textwidth]{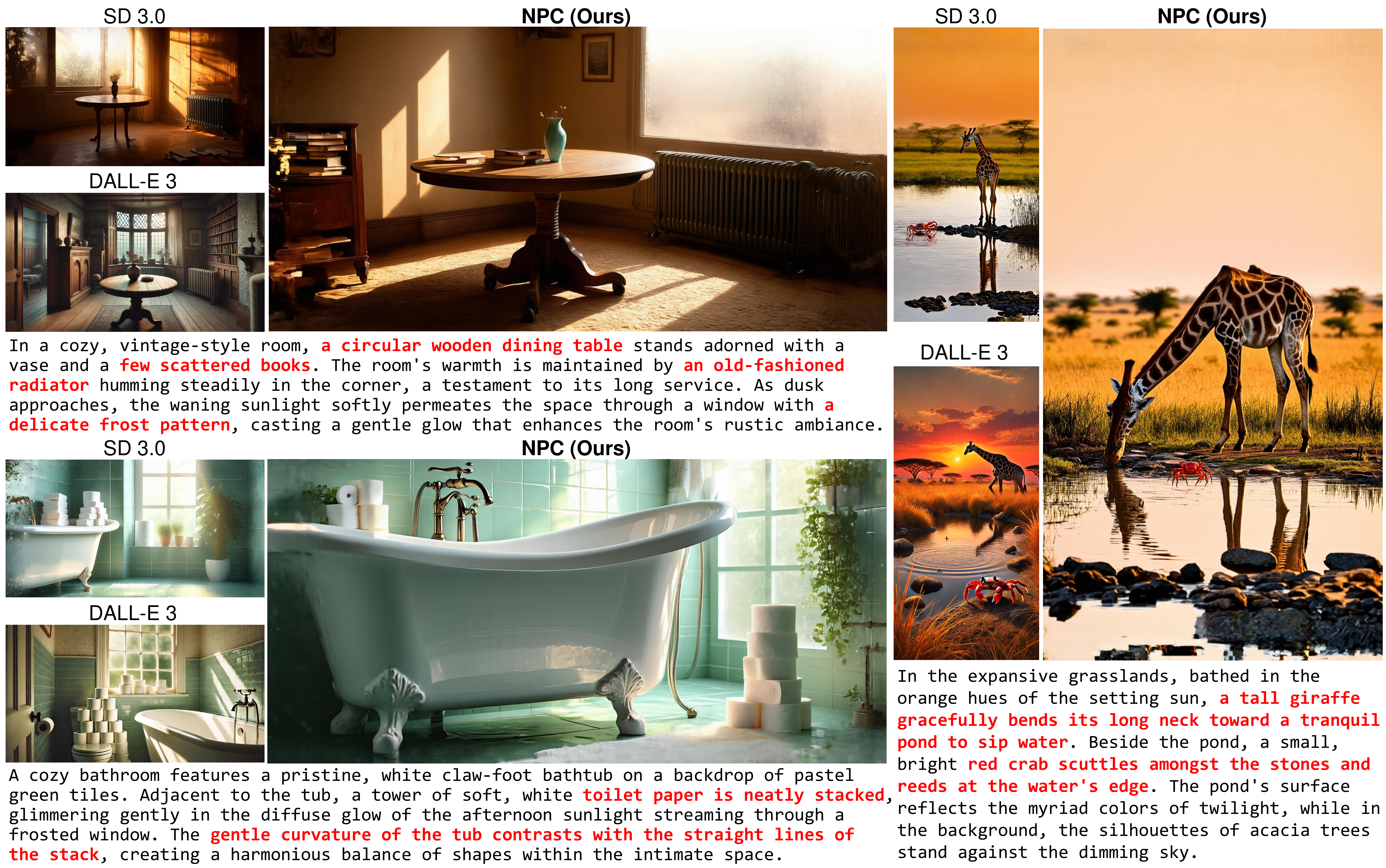}
  \end{center}
  \vskip -0.2in
  \caption{Examples using DPGBench, comparing NPC with SD3.0 and DALL-E 3, demonstrate that NPC is also effective for long, dense-context prompts. Quantitative comparison results are provided in the~\Cref{tab:dpg_bench}.}\label{fig:dpg}
\end{figure*}

\Cref{tab:dpg_bench} reports DPG-Bench results: NPC attains state-of-the-art relation accuracy (93.34) and the highest overall score, while remaining strong on entities. It lags top systems on global/other and is slightly behind on attributes, yet still captures fine-grained compositional details efficiently. \Cref{fig:dpg} further shows robust alignment in complex scenes.

\begin{table}
\centering
\begin{tabular}{lrrrr}
\toprule
 & pre-check & caption & propose & salient score \\
\midrule
Time (s) & 2.77 & 4.55 & 0.78 & 23.48 \\
\bottomrule
\end{tabular}
\vskip -0.1in
\caption{Computation time by stage.}\label{tab:computation}
\end{table}

\section{Experimental Details}
\paragraph{Software and Hardware} All experiments used PyTorch and an NVIDIA A40 GPU. 
\paragraph{Inference time and Memory}
Results averaged over 10 randomly sampled cases (as in \Cref{tab:computation}): pre-check about 2.77 s, caption about 4.55 s, proposal about 0.78 s, and scoring about 23.48 s. Scoring is the longest stage, but the overall increase remains marginal in practice. Memory changes were only marginal: we use the GPT API for captioning/proposal (no local GPU growth), and the salient-score stage ran on CPU, so GPU usage remained effectively unchanged.

\section{Additional Results for \Cref{sec:understanding}}\label{sec:additional_result}
For further validation of our analysis in \Cref{sec:understanding}, we compute salient attention scores on a broader set of prompts, as summarized in \Cref{tab:prompt-targeted-untargeted}. Salient tokens are manually selected from the generated images to pinpoint regions where text–image alignment fails; this manual curation is necessary to precisely measure how much attention increases on the misaligned parts, providing a more rigorous and reliable verification of our analysis. Across ten additional prompts, both the targeted and untargeted settings consistently yield higher salient attention scores than the base, reinforcing the robustness of our findings.

\begin{table*}[t]
\centering
\begin{adjustbox}{max width=\textwidth}
\begin{tabular}{lccccccc}
\toprule
\multirow{2}{*}{\textbf{Method}} & \multicolumn{3}{c}{\textbf{Attribute Binding}} & \multicolumn{2}{c}{\textbf{Object Relationship}} & \multirow{2}{*}{\textbf{Complex ↑}} & \multirow{2}{*}{\textbf{AVG ↑}} \\
\cmidrule(lr){2-4} \cmidrule(lr){5-6}
& \textbf{Color ↑} & \textbf{Shape ↑} & \textbf{Texture ↑} & \textbf{Spatial ↑} & \textbf{Non-Spatial ↑} & & \\
\midrule
DPO-Diff~\cite{pmlr-v235-wang24ar}  & 0.4405 & 0.4381 & 0.4949 & 0.1445 & 0.3182 & 0.4119 & 0.3746 \\ 
\rowcolor{blue!10}{\textbf{NPC$_{\text{SDXL}}$}}            & \textbf{0.7950} & \textbf{0.5681} & \textbf{0.6889} & \textbf{0.2545} & \textbf{0.3191} & \textbf{0.5171} & \textbf{0.5237} \\ 
\bottomrule

\end{tabular}
\end{adjustbox}
\caption{Comparison with negative prompt optimization method (\Cref{sec:npo}).}
\label{tab:npo}

\end{table*}

\section{Model-Agnostic Effectiveness of NPC}\label{sec:model_agnostic} 
To assess the generality of NPC beyond a single model or component choice, we evaluate its effectiveness across diverse generative architectures and MLLM or LLM configurations.
As shown in~\Cref{tab:sd-series}, across all three base models—SDXL~\cite{podellsdxl}, SD3~\cite{esser2024scaling}, and FLUX~\cite{black-forest-labs-no-date}—applying NPC consistently improves performance on every evaluated attribute. The gains are substantial and uniform, demonstrating that NPC is not model-specific but transfers effectively across architectures with different capacities. These results confirm the general robustness and broad applicability of NPC.

We further evaluate NPC using an alternative component stack in which the verifier and captioner are implemented with Qwen2.5-VL~\cite{bai2025qwen2} and the proposer with Qwen3~\cite{yang2025qwen3}. As shown in Table~\ref{tab:qwen-gpt}, the Qwen-based configuration achieves performance comparable to the GPT-based setup across all attributes, demonstrating that NPC remains effective even when substituting the underlying LLM components. This again highlights the model-agnostic nature of NPC and its robustness across different verifier, captioner, and proposer architectures.

\begin{figure}[t]
  \begin{center}
\includegraphics[width=\columnwidth]{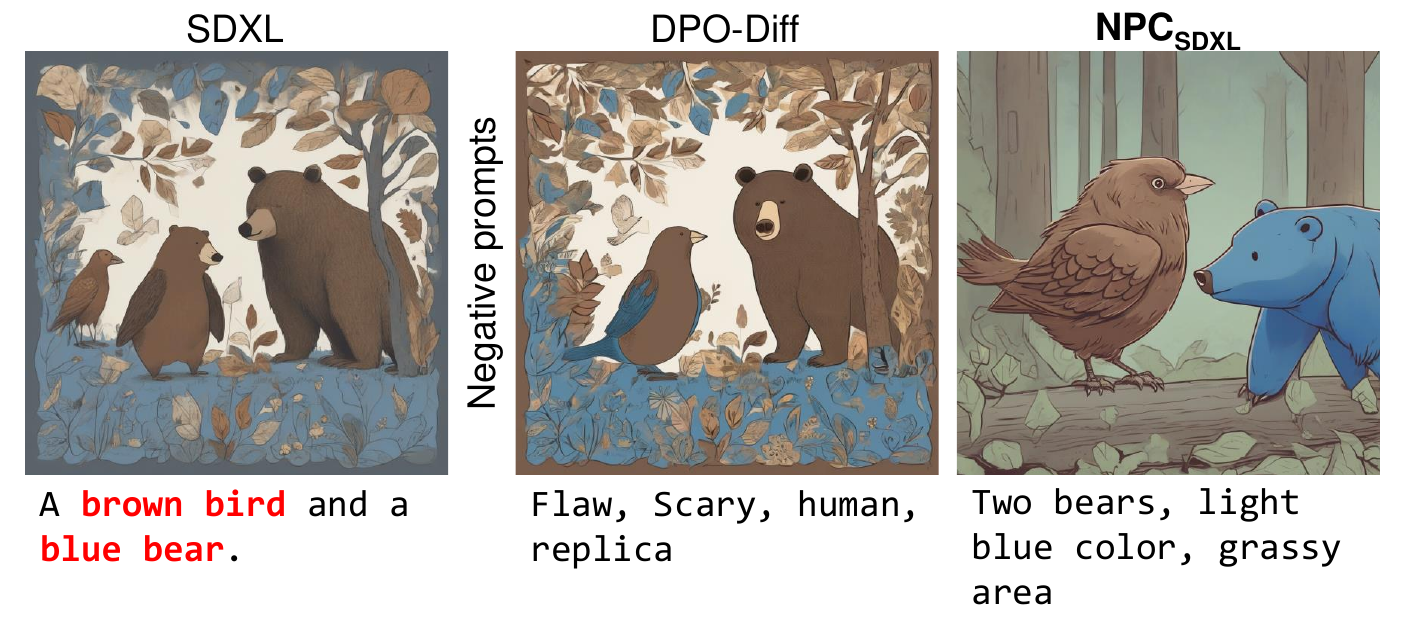}
  \end{center}
\vskip -0.2in
   \caption{Examples of DPO-Diff (\Cref{sec:npo}).}
   \label{fig:npo}
\end{figure}
\section{Comparison with Negative Prompt Optimization Method}\label{sec:npo} Many prompt optimization approaches for enhancing image quality have been explored~\cite{hao2023optimizing, Mo_2024_CVPR}. Since NPC can also be considered a negative prompt optimization method, we compare it with DPO-Diff~\cite{pmlr-v235-wang24ar}, a recently proposed approach that designs a compact space and employs shortcut gradient search to optimize negative prompts. As shown in~\Cref{tab:npo}, NPC outperforms DPO-Diff in T2I-CompBench since DPO-Diff is not specifically designed for alignment (examples of negative prompts from DPO-Diff are provided in~\Cref{fig:npo}). Moreover, unlike DPO-Diff, which requires training, NPC enhances alignment purely through inference, making it a more efficient alternative.

\section{LLM Prompts}\label{sec:llm_prompt}
In \Cref{tab:verifier}, \Cref{tab:captioner}, and \Cref{tab:proposer}, we provide the full LLM prompts used for the verifier, captioner, and proposer. \Cref{tab:prompt_ablation} shows the LLM prompt used in the ablation study of the prompt-only proposer in \Cref{sec:ablation_studies}.

\begin{table*}[h!]
\centering
\begin{tabular}{lccccccc}
\toprule
\multirow{2}{*}{\textbf{Model}} & \multicolumn{3}{c}{\textbf{Attribute Binding}} & \multicolumn{2}{c}{\textbf{Object Relationship}} & \multirow{2}{*}{\textbf{Complex ↑}} & \multirow{2}{*}{\textbf{AVG ↑}} \\
\cmidrule(lr){2-4} \cmidrule(lr){5-6}
& \textbf{Color ↑} & \textbf{Shape ↑} & \textbf{Texture ↑} & \textbf{Spatial ↑} & \textbf{Non-Spatial ↑} & & \\
\midrule
Stable Diffusion 2 & 0.5065 & 0.4221 & 0.4922 & 0.1342 & 0.3096 & 0.3386 & 0.3672 \\
SDXL & 0.6369 & \underline{0.5408} & 0.5637 & 0.2032 & 0.3110 & 0.4091 & 0.4441 \\
\midrule

\tikzcircle[red, fill=red, opacity=0.5]{0.08cm}Attn-Exct v2 & 0.6400 & 0.4517 & 0.5963 & 0.1455 & 0.3109 & 0.3401 & 0.4140 \\ 
\tikzcircle[red, fill=red, opacity=0.5]{0.08cm}SynGen         & 0.7000    & 0.4550   & 0.6010    & 0.2260   & 0.3100         & 0.3339    & 0.4376       \\
\tikzcircle[blue, fill=blue, opacity=0.5]{0.08cm}LMD           & 0.4814   & 0.4865   & 0.5699   & 0.2537   & 0.2828          & 0.3323    &   0.4011    \\
\tikzcircle[blue, fill=blue, opacity=0.5]{0.08cm}InstanceDiffusion & 0.5433   & 0.4472   & 0.5293   & \textbf{0.2791}   & 0.2947         & 0.3602    & 0.4089       \\
\tikzcircle[green, fill=green, opacity=0.5]{0.08cm}GORS & 0.6603 & 0.4785 & 0.6287 & 0.1815 & \textbf{0.3193} & 0.3328 & 0.4335\\ 
\tikzcircle[green, fill=green, opacity=0.5]{0.08cm}CoMat            & \underline{0.7827}   & 0.5329   & \underline{0.6468}   & 0.2428   & 0.3187         & \underline{0.3680}    & \underline{0.4819}       \\

\textbf{NPC$_{\text{SDXL}}$ (Ours)}              & \textbf{0.7950} & \textbf{0.5681} & \textbf{0.6889} & \underline{0.2545} & \underline{0.3191}       & \textbf{0.5171}  & \textbf{0.5237} \\
\bottomrule
\end{tabular}
\caption{T2I-CompBench. Comparison of various methods across different attributes. The best performances are highlighted in bold and the second-best are underlined. The red circle (\tikzcircle[red, fill=red, opacity=0.5]{0.08cm}) represents attention-map-based approaches, the blue circle (\tikzcircle[blue, fill=blue, opacity=0.5]{0.08cm}) indicates layout-based approaches, and the green circle (\tikzcircle[green, fill=green, opacity=0.5]{0.08cm}) denotes feedback-based approaches. Performance for the baselines is based on reported values.}
\label{tab:maintable}
\end{table*}
\begin{table*}[t]
\centering
\setlength{\tabcolsep}{8pt}
\renewcommand{\arraystretch}{1.2}
\begin{tabular}{lrrrrrr}
\toprule
\textbf{Method} & \textbf{Global}$\uparrow$ & \textbf{Entity}$\uparrow$ & \textbf{Attribute}$\uparrow$ & \textbf{Relation}$\uparrow$ & \textbf{Other}$\uparrow$ & \textbf{Overall}$\uparrow$ \\
\midrule
SDXL~\cite{podellsdxl}                 & 83.27 & 82.43 & 80.91 & 86.76 & 80.41 & 74.65 \\
Hunyuan-DiT~\cite{li2024hunyuan}       & 84.59 & 80.59 & 88.01 & 74.36 & 86.41 & 78.87 \\
DALLE3~\cite{betker2023improving}      & \underline{90.97} & 89.61 & 88.39 & 90.58 & \underline{89.83} & 83.50 \\
SD3-medium~\cite{esser2024scaling}     & 87.90 & \underline{91.01} & 88.83 & 80.70 & 88.68 & 84.08 \\
FLUX.1-dev~\cite{black-forest-labs-no-date} & 82.10 & 89.50 & 88.70 & \underline{91.10} & 89.40 & 84.00 \\
OmniGen~\cite{xiao2025omnigen}         & 87.90 & 88.97 & 88.47 & 87.95 & 83.56 & 81.16 \\
\midrule
Show-o~\cite{xie2025showo}             & 79.33 & 75.44 & 78.02 & 84.45 & 60.80 & 67.27 \\
EMU3~\cite{wang2024emu3}               & 85.21 & 86.68 & 86.84 & 90.22 & 83.15 & 80.60 \\
TokenFlow-XL~\cite{qu2025tokenflow}    & 78.72 & 79.22 & 81.29 & 85.22 & 71.20 & 73.38 \\
Janus Pro~\cite{chen2025janus}         & 86.90 & 88.90 & 89.40 & 89.32 & 89.48 & 84.19 \\
T2I-R1~\cite{jiang2025t2i}             & \textbf{91.79} & 90.23 & 89.05 & 90.13 & 89.48 & 84.76 \\
UniWorld-V1~\cite{lin2025uniworld}     & 83.64 & 88.39 & 88.44 & 89.27 & 87.22 & 81.38 \\
OmniGen2~\cite{wu2025omnigen2}         & 88.81 & 88.83 & \underline{90.18} & 89.37 & \textbf{90.27} & 83.57 \\
BAGEL~\cite{deng2025emerging}          & 88.94 & 90.37 & \textbf{91.29} & 90.82 & 88.67 & \underline{85.07} \\
\rowcolor{blue!10}
\rowcolor{blue!10}
\textbf{NPC$_{\text{FLUX}}$}            & 83.88 & \textbf{91.12} & 87.69 & \textbf{93.34} & 83.03 & \textbf{85.38} \\
\bottomrule
\end{tabular}
\caption{DPG-Bench comparison (higher is better, $\uparrow$). Bold indicates the best value in each column; \underline{underline} indicates the second-best.}
\label{tab:dpg_bench}
\end{table*}


\newcolumntype{Y}{>{\RaggedRight\arraybackslash}X}
\newcommand{\gain}[1]{\cellcolor{green!12}\textbf{#1}}

\begin{table*}[t]
\centering
\small
\setlength{\tabcolsep}{4pt}
\renewcommand{\arraystretch}{1.15}
\begin{tabularx}{\textwidth}{Y l Y Y r r r}
\toprule
\textbf{Prompt} & \textbf{Salient} & \textbf{Targeted} & \textbf{Untargeted} & \textbf{Base}$\uparrow$ & \textbf{Targeted}$\uparrow$ & \textbf{Untargeted}$\uparrow$ \\
\midrule
A photo of two trains, three bottles, and two tennis rackets & three & 2 bottles & toy stream locomotives & 0.0831 & \gain{0.1018} & \gain{0.0839} \\
\multicolumn{5}{r}{} & \scriptsize +0.0187 & \scriptsize +0.0008 \\
\addlinespace[2pt]
A photo of three bowls and two apples & three & stacked bowls & shiny apples & 0.0746 & \gain{0.1018} & \gain{0.0860} \\
\multicolumn{5}{r}{} & \scriptsize +0.0272 & \scriptsize +0.0114 \\
\addlinespace[2pt]
A photo of two teddy bears and three laptops & three & one laptop & wooden table & 0.0583 & \gain{0.0642} & \gain{0.0703} \\
\multicolumn{5}{r}{} & \scriptsize +0.0059 & \scriptsize +0.0120 \\
\addlinespace[2pt]
A photo of a purple chair, a blue bottle, and a white vase & white & purple vase & bulbous object & 0.0427 & \gain{0.0512} & \gain{0.0473} \\
\multicolumn{5}{r}{} & \scriptsize +0.0085 & \scriptsize +0.0046 \\
\addlinespace[2pt]
A photo of a larger donut on the above and a smaller zebra on the below & zebra & two donuts & colorful sprinkles & 0.0344 & \gain{0.0371} & \gain{0.0428} \\
\multicolumn{5}{r}{} & \scriptsize +0.0027 & \scriptsize +0.0084 \\
\addlinespace[2pt]
A photo of seven bears & seven & six bears & wildflowers & 0.0926 & \gain{0.1031} & \gain{0.1070} \\
\multicolumn{5}{r}{} & \scriptsize +0.0105 & \scriptsize +0.0144 \\
\addlinespace[2pt]
A photo of six vases & six & seven vases & glass vases & 0.0863 & \gain{0.0947} & \gain{0.0963} \\
\multicolumn{5}{r}{} & \scriptsize +0.0084 & \scriptsize +0.0100 \\
\addlinespace[2pt]
A photo of three dining tables and three horses & three & two horses & chandelier & 0.1571 & \gain{0.1638} & \gain{0.1630} \\
\multicolumn{5}{r}{} & \scriptsize +0.0067 & \scriptsize +0.0059 \\
\addlinespace[2pt]
A photo of two stop signs and one birds & two & one stop sign & television sets & 0.0790 & \gain{0.0850} & \gain{0.0820} \\
\multicolumn{5}{r}{} & \scriptsize +0.0060 & \scriptsize +0.0030 \\
\addlinespace[2pt]
A photo of two green tvs and two pink potted plant & two & one green TV & antennas & 0.1178 & \gain{0.1464} & \gain{0.1289} \\
\multicolumn{5}{r}{} & \scriptsize +0.0286 & \scriptsize +0.0111 \\
\bottomrule
\end{tabularx}
\caption{Additional results for \Cref{sec:understanding}. results (higher is better, $\uparrow$). In each pair of rows, the first row lists \textit{Base}, \textit{Targeted}, and \textit{Untargeted} scores (cells in \textbf{green, bold} exceed \textit{Base}); the second row shows unlabeled absolute improvements in the last two columns \emph{(Targeted$-$Base,\ Untargeted$-$Base)}. All prompts show \textit{Targeted} and \textit{Untargeted} $>$ \textit{Base}.}
\label{tab:prompt-targeted-untargeted}
\end{table*}

\begin{table*}[t]
\centering
\setlength{\tabcolsep}{8pt}
\begin{tabular}{lcccccccc}
\toprule
\textbf{Model} & \textbf{Color} & \textbf{Count} & \textbf{Color/Count} & \textbf{Color/Pos} & \textbf{Pos/Count} & \textbf{Pos/Size} & \textbf{Multi-Count} & \textbf{Overall}\\
\midrule
SDXL~\cite{podellsdxl}              
      & 0.050 & 0.375 & 0.000 & 0.000 & 0.000 & 0.000 & 0.000 & 0.061 \\
NPC\textsubscript{SDXL}             
      & \textbf{0.500} & \textbf{0.575} & \textbf{0.200} & \textbf{0.400} 
      & \textbf{0.325} & \textbf{0.425} & \textbf{0.500} & \textbf{0.417} \\
\midrule
SD3~\cite{esser2024scaling} 
      & 0.550 & 0.500 & 0.125 & 0.350 & 0.175 & 0.150 & 0.225 & 0.296 \\
NPC\textsubscript{SD3}               
      & \textbf{0.650} & \textbf{0.675} & \textbf{0.325} & \textbf{0.550} 
      & \textbf{0.575} & \textbf{0.600} & \textbf{0.625} & \textbf{0.571} \\
\midrule
FLUX~\cite{black-forest-labs-no-date}
      & 0.350 & 0.625 & 0.150 & 0.275 & 0.200 & 0.375 & 0.225 & 0.314 \\
NPC\textsubscript{FLUX}               
      & \textbf{0.550} & \textbf{0.675} & \textbf{0.350} & \textbf{0.525} 
      & \textbf{0.550} & \textbf{0.725} & \textbf{0.625} & \textbf{0.571} \\
\bottomrule
\end{tabular}
\caption{Comparison of metric scores across three models before and after applying NPC. In all cases, NPC improves performance across all evaluated attributes.}
\label{tab:sd-series}
\end{table*}

\begin{table*}[t]
\centering
\setlength{\tabcolsep}{8pt}
\begin{tabular}{lcccccccc}
\toprule
\textbf{Method} & \textbf{Color} & \textbf{Count} & \textbf{Color/Count} & \textbf{Color/Pos} & \textbf{Pos/Count} & \textbf{Pos/Size} & \textbf{Multi-Count} & \textbf{Overall} \\
\midrule
Qwen              
      & 0.650 & 0.575 & 0.400 & 0.500 
      & 0.575 & 0.650 & 0.600 & 0.564 \\
GPT               
      & 0.550 & 0.675 & 0.350 & 0.525 
      & 0.550 & 0.725 & 0.625 & 0.571 \\
\bottomrule
\end{tabular}
\caption{Comparison of metric scores between Qwen-based and GPT-based components for NPC.}
\label{tab:qwen-gpt}
\end{table*}

\begin{table*}[t]
\centering
\small
\begin{tabularx}{\textwidth}{>{\ttfamily\arraybackslash}X}
\toprule
<System Prompt>: You are an expert image evaluator. \\
\\
Decide if a single candidate image fully satisfies a natural-language INSTRUCTION and, if provided, an expectation CHECKLIST. \\
\\
Strict rules: \\
1) All expectations must be satisfied: \\
\ \ \ - Object classes, counts, colors, spatial relations (above/below/left/right by pixel position), size/relative scale. \\
2) The image must be a single coherent, natural, photo-like scene. \\
\ \ \ - Reject stylized (cartoons, sketches, anime), collages, multi-panels, or text-only. \\
3) Be strict and conservative. If uncertain, mark incorrect. \\
\\
Return a strict JSON object only: \\
\{"correct\textquotedblright: 1 or 0, "score\textquotedblright: float in [0,1], "reason\textquotedblright: "brief explanation\textquotedblright\} \\
No extra text. "score\textquotedblright\ should reflect how well the image meets the INSTRUCTION (and CHECKLIST), even if correct=0. \\
\\
<User Prompt>: POSITIVE PROMPT: \\
\$positive\_prompt \\
\\
CHECKLIST (optional): \\
\$checklist \\
\\
Judge the attached image against the INSTRUCTION (and CHECKLIST if present). \\
Return exactly: \\
\{"correct\textquotedblright: 1 or 0, "score\textquotedblright: float in [0,1], "reason\textquotedblright: "brief explanation\textquotedblright\} \\
\\
\bottomrule
\end{tabularx}
\caption{LLM prompt for \emph{Verifier}.}
\label{tab:verifier}
\end{table*}

\begin{table*}[t]
\centering
\small
\begin{tabularx}{\textwidth}{>{\ttfamily\arraybackslash}X}
\toprule
<System Prompt>: You are a meticulous visual describer. \\
Write a comprehensive caption (90--160 words) describing concrete, falsifiable details only. \\
Cover: main objects (counts, colors, materials), textures/patterns, spatial relations, background elements, \\
lighting, style/mood, any visible text. \\
Avoid speculation, emotions, and camera metadata. Output plain text only. \\
\\
<User Prompt>: Describe this image. \\
\bottomrule
\end{tabularx}
\caption{LLM prompt for \emph{Captioner}.}
\label{tab:captioner}
\end{table*}


\begin{table*}[t]
\centering
\small
\begin{tabularx}{\textwidth}{>{\ttfamily\arraybackslash}X}
\toprule
<System Prompt>: You are a precise prompt editor. \\
From a given POSITIVE PROMPT and an image CAPTION, extract concise negative prompt candidates (1--6 words, all lowercase) that would most effectively steer generation away from elements that contradict the POSITIVE PROMPT. \\
Return a strict JSON object as specified. Do not include explanations. \\
\\
<User Prompt>: POSITIVE PROMPT: \\
\$positive\_prompt \\
\\
CAPTION: \\
\$caption \\
\\
GENERAL\_FALLBACKS (use only if needed to reach K): \\
\$fallbacks \\
\\
K (max total candidates): \$k \\
\\
Task \\
1) From the CAPTION, find elements that are NOT requested by the POSITIVE PROMPT (consider synonyms, inflections, paraphrases). \\
2) Prioritize contradictions to the POSITIVE PROMPT (object identity, attributes like color/material/count/shape, relations/pose), and if contradictions are weak or absent, pick items at least tightly related to the POSITIVE PROMPT's objects (foreground-first). \\
3) Produce a flat list "candidates" of up to K concise negative phrases. \\
4) If you have fewer than K items, supplement with the most relevant items from GENERAL\_FALLBACKS while still obeying all constraints and avoiding anything implied by the POSITIVE PROMPT. \\
5) Return exactly K items if possible (use fallbacks to fill). Only return fewer than K if no valid item remains. \\
\\
Constraints \\
- Use ONLY elements explicitly present in the CAPTION. \\
- Exclude anything present in (or synonymous with) the POSITIVE PROMPT. \\
- Phrases: lowercase, 1--6 words, no "no/without/not", no quotes, no trailing punctuation, no duplicates. \\
\\
Return exactly: \\
\{"candidates": ["...", "...", "..."]\} \\
\bottomrule
\end{tabularx}
\caption{LLM prompt for \emph{Proposer}.}
\label{tab:proposer}
\end{table*}

\begin{table*}[t]
\centering
\small
\begin{tabularx}{\textwidth}{>{\ttfamily\arraybackslash}X}
\toprule
<System Prompt>: You are a precise prompt editor.\\
Given only a POSITIVE PROMPT (no caption) and K, produce up to K concise negative prompt candidates (1--6 words, all lowercase) that would most effectively steer image generation away from elements that contradict the POSITIVE PROMPT.\\
Return only strict JSON as specified. Do not include explanations.\\[0.6em]

\textbf{Task}\\
1) Parse the POSITIVE PROMPT to identify primary subjects/objects, key attributes (color/material/count/shape/pose), relations, scene/setting/time/lighting, viewpoint/composition, and medium/style/era.\\
2) Generate concise negative phrases that capture the most likely contradictions or near-misses to those elements, prioritized in this order: foreground object identity $\rightarrow$ attributes $\rightarrow$ relations/pose $\rightarrow$ composition/viewpoint $\rightarrow$ style/medium/era.\\
3) If strong contradictions are scarce, choose tightly related alternatives for the same foreground objects (e.g., sibling species, alternate colors/counts, opposite angles).\\[0.6em]

\textbf{Constraints}\\
-- Phrases: lowercase, 1--6 words, no ``no/without/not'', no quotes, no trailing punctuation, no duplicates.\\
-- Exclude anything present in (or synonymous with) the POSITIVE PROMPT.\\
-- Use only concepts reasonably inferred from the POSITIVE PROMPT’s domain; avoid unrelated items.\\
-- Foreground-first: prefer conflicts about the main subject before background or style.\\[0.6em]

<User Prompt>: POSITIVE PROMPT: \$positive\_prompt\\
K (max total candidates): \$k\\[0.4em]

Task:\\
Propose \$k negative prompt tokens likely to reduce undesirable or off-topic elements that could appear when generating this scene.\\[0.4em]

Return exactly:\\
\{"candidates": ["...", " ..."\}\\
\bottomrule
\end{tabularx}
\caption{LLM prompt for prompt ablation in Section 5.3.}
\label{tab:prompt_ablation}
\end{table*}

\end{document}